\documentclass{article}
\pdfoutput=1

\usepackage{PRIMEarxiv}

\usepackage[utf8]{inputenc} 
\usepackage[T1]{fontenc}    
\usepackage{hyperref}       
\usepackage{url}           
\usepackage{booktabs}       
\usepackage{amsfonts}      
\usepackage{nicefrac}       
\usepackage{microtype}      
\usepackage{lipsum}
\usepackage{fancyhdr}       
\usepackage{graphicx}       
\graphicspath{{media/}}     
\usepackage{multirow} 
\usepackage{bbding} 
\usepackage{caption} 
\usepackage{subcaption} 
	
\usepackage{xcolor}
\title{Multi Modal Facial Expression Recognition with Transformer-Based Fusion Networks and Dynamic Sampling}

\author{
  JUN-HWA KIM \\
  Department of Electronics and Electrical Engineering \\
  Dongguk University \\
  Seoul, Korea\\
  \texttt{jhkim414@dongguk.edu} \\
    \And
  NAMHO KIM \\
  Department of Electronics and Electrical Engineering \\
  Dongguk University \\
  Seoul, Korea\\
  \texttt{namho96@dgu.ac.kr} \\
   \And
  CHEE SUN WON \\
  Department of Electronics and Electrical Engineering \\
  Dongguk University \\
  Seoul, Korea\\
  \texttt{cswon@dongguk.edu} \\
}

\begin{document}
\maketitle

\begin{abstract}
Facial expression recognition is an essential task for various applications, including emotion detection, mental health analysis, and human-machine interactions. In this paper, we propose a multi-modal facial expression recognition method that exploits audio information along with facial images to provide a crucial clue to differentiate some ambiguous facial expressions. Specifically, we introduce a Modal Fusion Module (MFM) to fuse audio-visual information, where image and audio features are extracted from Swin Transformer. Additionally, we tackle the imbalance problem in the dataset by employing dynamic data resampling. Our model has been evaluated in the Affective Behavior in-the-wild (ABAW) challenge of CVPR 2023.
\end{abstract}


\section{Introduction}

Facial Expression Recognition (FER) has received much attention in recent years due to the increasing demands in various applications such as emotion detection, mental health analysis,  and human-machine interactions \cite{muhammad2017facial,davoudi2019intelligent}. With the advent of deep learning architectures, the performance of facial expression recognition has improved significantly. Note that the previous methods rely mostly on the image features to recognize the facial expressions. However, some facial expressions can be more clearly recognized by checking the audio on top of the image data \cite{zhang2022transformer,meng2022valence}. That is, some facial expressions are associated with vocal and auditory cues such as tone of voice, rate of speech, and loudness. So, they can provide crucial information about a certain emotional status. To address this issue, we propose an approach for facial expression recognition that fuses image and audio features through a Multi Fusion Module (MFM). The MFM is designed to adaptively learn the importance of each modality in the fusion process. Each image and audio feature is extracted using a Swin Transformer \cite{liu2021swin}. 

To evaluate our proposed approach, we conduct experiments on the Aff-Wild2 dataset \cite{kour2014real, kour2014fast,hadash2018estimate,kollias2022abaw, kollias2021analysing, kollias2020analysing, kollias2021distribution, kollias2021affect, kollias2019expression, kollias2019face, kollias2019deep, zafeiriou2017aff, kollias2023abaw, kollias2023abaw2}. The Aff-Wild2 dataset is a large-scale in-the-wild dataset for the 5th Affective Behavior Analysis in-the-wild (ABAW 2023) competition in conjunction with CVPR 2023.  

\section{Methodology}

\subsection{Multi Modal Fusion Network}

\begin{figure}[h]
    \includegraphics[width=15cm]{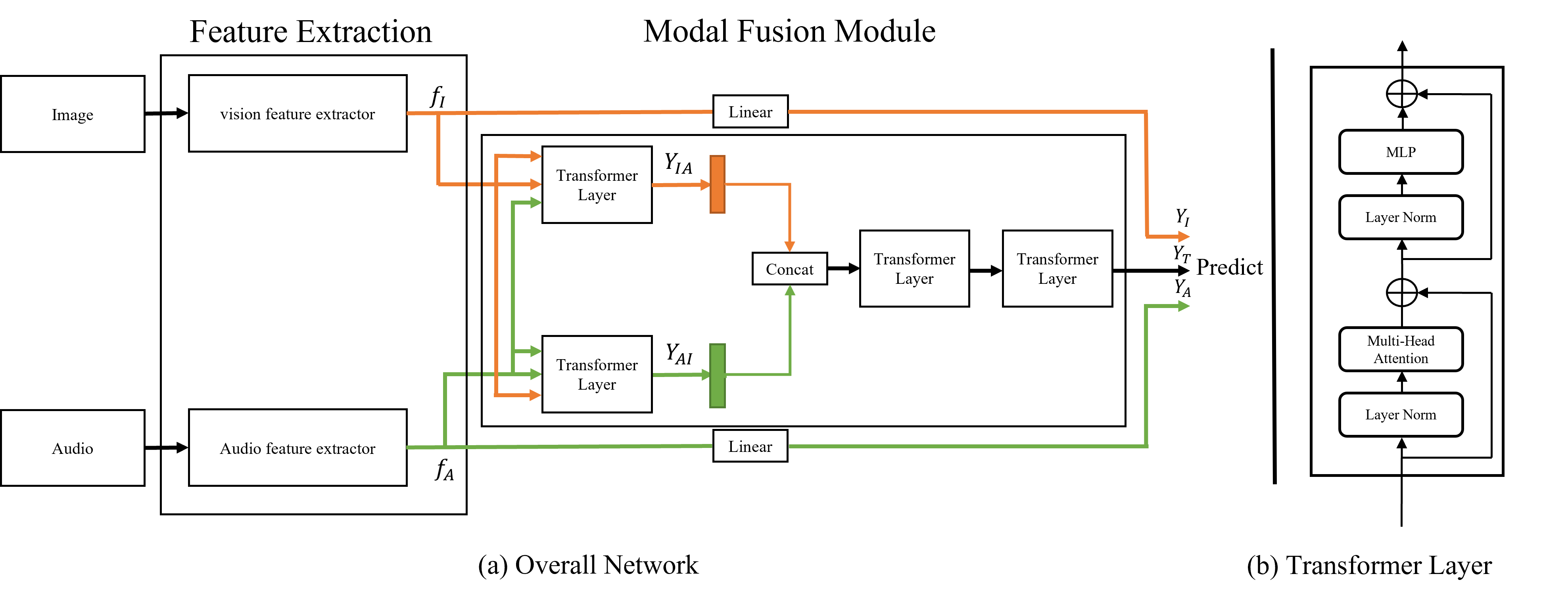}
    \caption{Overall structure of the proposed network.}
    \label{network}
\end{figure}

As shown in Figure \ref{network}, our facial expression recognition with a modal fusion network involves four main steps. Firstly, the image and audio data are trained separately using a Swin Transformer \cite{liu2021swin} to obtain features of $f_{I}$ and $f_{A}$, respectively. Next, these features are fed into  Modal Fusion Module (MFM). The MFM fuse them through the process of co-attention with the Transformer Layers. Specifically, we forward $f_{A}$ for Query and $f_{I}$ for Key and Value to fuse audio information based on an image. Similarly, forward $f_{I}$ for Query and $f_{A}$ for Key and Value to fuse image information based on audio. The fused attention output vector $Y_{IA}$ and $Y_{AI}$ can be represented as follows:

\begin{equation}
    Y_{IA} = Attention(f_{A}, f_{I}, f_{I}) = softmax(\frac{f_{A}f_{I}^{T}}{\sqrt{d_{k}}})f_{I},
\end{equation}

\begin{equation}
    Y_{AI} = Attention(f_{I}, f_{A}, f_{A}) = softmax(\frac{f_{I}f_{A}^{T}}{\sqrt{d_{k}}})f_{A},
\end{equation}
where $d_{k}$ is the dimensionality of the key vector and $T$ denotes the Transpose.

\begin{equation}
    Y = Concat(Y_{IA}, Y_{AI}),
\end{equation}

where $Y$ denotes the Fused two modalities. Then, $Y$ is passed through two transformer layers to obtain $Y_{T}$. Thirdly, image and audio features are separately fed through linear layers to obtain the output vectors $Y_{I}$ and $Y_{A}$, respectively. Finally, the three outputs, $Y_{I}$, $Y_{T}$, and $Y_{A}$, are ensembled to obtain the final prediction. The transformer layer uses the same structure as the encoder of ViT \cite{dosovitskiy2020image}.

\subsection{Dynamic sampling and Resampling}

\begin{figure}
    \includegraphics[width=15cm]{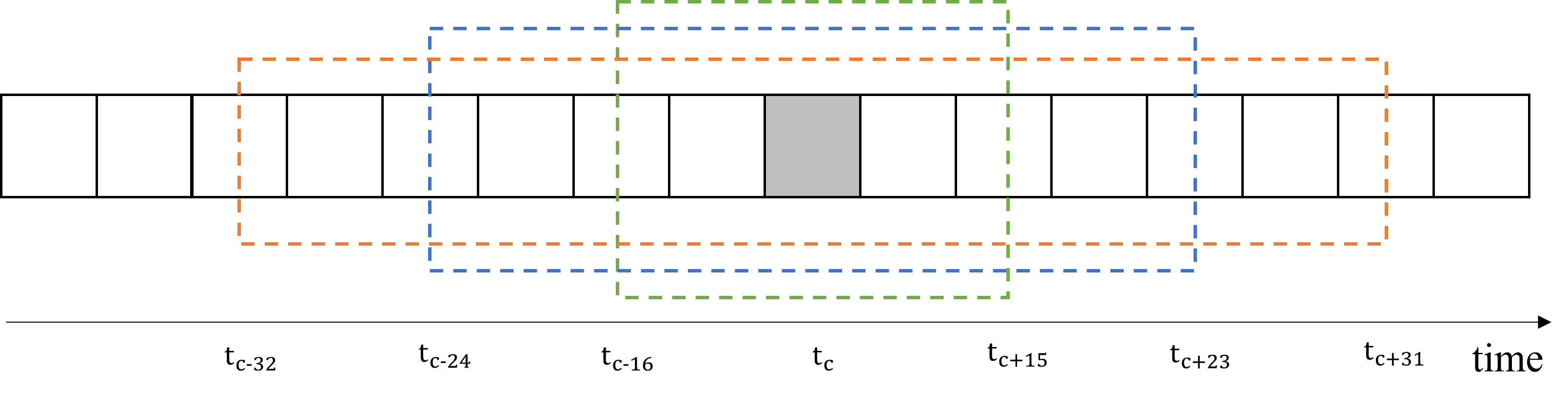}
    \caption{Dynamic sampling with three different windows.}
    \label{dynamic}
\end{figure}

Facial expressions may occur abruptly from one frame to the next one, but there are usually strong influences among adjacent frames. To cover the different ranges of facial expressions, the audio data was sampled in a dynamic manner. As shown in Figure \ref{dynamic}, to capture the temporal dynamics of facial expressions, we set three different lengths of windows centered on the current frame $t_{c}$, extracting three dynamic audio features. The ranges of the three dynamic windows are $t_{s} = \{t_{c-16}, ..., t_{c}, ... t_{c+15}\}$, $t_{m} = \{t_{c-24}, ..., t_{c}, ... t_{c+23}\}$, and $t_{l} = \{t_{c-32}, ..., t_{c}, ... t_{c+31}\}$. Having adjacent frames in the window, included frames may have different ground truth labels from each other. Therefore, it is necessary to use only the windows in which the majority of ground truth label in the window is identical to that of the center frame. To sort out such windows we set the thresholding values for the three windows as 0.8 for $t_{s}$, 0.65 for $t_{m}$, and 0.5 for $t_{l}$. So, for example, if less than $80\%$ (i.e., the threshold $0.8$) of the frames in the window $t_{s}$ have the identical ground truth label with that of the center frame, then we exclude this $t_s$ from the training. 

Table \ref{table_dataset_stat} shows the number and ratio of images for each class in the Aff-wild2 dataset \cite{kollias2023abaw}. Note that due to severe distribution imbalance among different classes as in Table \ref{table_dataset_stat}, there is a high risk of overfitting to the majority class if trained by class-imbalanced data. To keep the class balance, we make a subsampling for the majority class in the training phase. While all training data in the minority classes are used for the training, not all training data in the majority class are used. That is, the training data in the majority class such as `Happiness' are subsampled for the training. 

\begin{table*}[h]
  \caption{Number of image distribution among expression classes for the Aff-Wild2 dataset.}
  \centering
  \begin{tabular}{ccccccccccc}
    \toprule
    \multicolumn{2}{c}{\multirow{2}{*}{Dataset}} & \multicolumn{8}{c}{Expression} & \multirow{2}{*}{Total} \\ \cmidrule{3-10}
    {} & {} & Neutral & Anger & Disgust & Fear & Happiness & Sadness & Surprise & Other & {} \\
    
    \midrule
    \multirow{4}{*}{Train} & Number & \multirow{2}{*}{180227} & \multirow{2}{*}{17153} & \multirow{2}{*}{10978} & \multirow{2}{*}{9110} & \multirow{2}{*}{97302} & \multirow{2}{*}{80671} & \multirow{2}{*}{32033} & \multirow{2}{*}{169887} & \multirow{2}{*}{597361}\\ 
    &  of Data & \\
    \cmidrule{2-11}
    &  Ratio to & \multirow{2}{*}{0.302} & \multirow{2}{*}{0.029} & \multirow{2}{*}{0.018} & \multirow{2}{*}{0.015} & \multirow{2}{*}{0.163} & \multirow{2}{*}{0.135} & \multirow{2}{*}{0.053} & \multirow{2}{*}{0.284} & \multirow{2}{*}{1.000}\\
    &  total & \\
    \midrule
    \multirow{4}{*}{Validation} & Number & \multirow{2}{*}{83176} & \multirow{2}{*}{6127} & \multirow{2}{*}{5322} & \multirow{2}{*}{8473} & \multirow{2}{*}{34941} & \multirow{2}{*}{25594} & \multirow{2}{*}{12338} & \multirow{2}{*}{108259} & \multirow{2}{*}{284230} \\
    &  of Data & \\
    \cmidrule{2-11}
    &  Ratio to & \multirow{2}{*}{0.293} & \multirow{2}{*}{0.022} & \multirow{2}{*}{0.019} & \multirow{2}{*}{0.030} & \multirow{2}{*}{0.123} & \multirow{2}{*}{0.090} & \multirow{2}{*}{0.043} & \multirow{2}{*}{0.381} & \multirow{2}{*}{1.000} \\
    &  total & \\
    \bottomrule
  \end{tabular}
  
  \label{table_dataset_stat}
\end{table*}

\section{Experiment results}

\subsection{Implementation details}

Our experiments were conducted in the PyTorch environment on a desktop computer with the following specifications: Ubuntu 20.4 operating system, 32GB RAM, and GPU GeForce RTX 1080ti with 11GB memory. We used the Aff-Wild2 dataset \cite{kollias2023abaw} for the experiments, which consists of 8 classes, including 7 emotions ('Neutral', 'Anger', 'Disgust', 'Fear', 'Happiness', 'Sadness', 'Surprise') and the 'Other' class. Table \ref{table_dataset_stat} shows the number of images and imbalance ratios for each of the eight emotions in Aff-Wild2. For preprocessing the Aff-Wild2 dataset, we used only cropped aligned images provided by the competition organizers and resized them to $224\times 224$ for the network input. The evaluation dataset was also provided by the competition organizers.

We adopted Tiny model of Swin Transformer \cite{liu2021swin} for the feature extraction network. Audio spectrograms were generated by using a log mel spectrogram with the following parameters: 2048 FFT points, a hop length of 1024 samples, and 128 frequency bins.

\subsection{Evaluation metric}

In the Affective Behavior Analysis in-the-wild (ABAW) Expression Classification competition, F1 score is used as an evaluation metric to assess the performance. F1 score is a commonly used metric that combines precision and recall into a single measure, and it is particularly useful in imbalanced datasets where the number of samples in each class is different. 
The $F_1$ score is defined as 
\begin{equation}
    F_{1} = \frac{1}{n}\sum_{i}^{n}{\frac{2 \times precision_{i} \times recall_{i}}{precision_{i} + recall_{i}}},
\end{equation}
where $n$ is the number of emotion class, $precision_{i}$ is the precision of the $i$-th class, and $recall_{i}$ is the recall of the $i$-th class.

\subsection{Results}

Table \ref{table_results} shows the results of our experiments on the validation set of Aff-Wild2 \cite{kollias2023abaw}. 

\begin{table}[h]

  \caption{Experimental results for Aff-Wild2 validation set.}
  \centering
  \begin{tabular}{cccccc}
    \toprule
    Model & Modal   & F1-score \\ \midrule
    Baseline & - & 23.00 \\
    \midrule
    \midrule
    Swin-Tiny & Image($Y_{I}$)  & 25.61  \\ 
    \midrule
    Swin-Tiny & Audio($Y_{A}$)  & 14.12 \\ 
    \midrule
    Swin-Tiny & Fusion($Y_{T}$)  & 25.09 \\ 
    \midrule
    Swin-Tiny & Image($Y_{I}$), Audio($Y_{A}$), Fusion($Y_{T}$)   & 27.77  \\

    \bottomrule
  \end{tabular}
  
  \label{table_results}
\end{table}

\section{Conclusion}

In this paper, we have proposed a network architecture that can effectively fuse the audio as well as the image features for the facial expression recognition problem. To this end we adopted Modal Fusion Module (MFM) with the transformer layers. Our model has been evaluated in the ABAW challenge with the Aff-Wild2 dataset.


\begin{thebibliography}{10}

\bibitem{muhammad2017facial}
Ghulam Muhammad, Mansour Alsulaiman, Syed~Umar Amin, Ahmed Ghoneim, and
  Mohammed~F Alhamid.
\newblock A facial-expression monitoring system for improved healthcare in
  smart cities.
\newblock {\em IEEE Access}, 5:10871--10881, 2017.

\bibitem{davoudi2019intelligent}
Anis Davoudi, Kumar~Rohit Malhotra, Benjamin Shickel, Scott Siegel, Seth
  Williams, Matthew Ruppert, Emel Bihorac, Tezcan Ozrazgat-Baslanti, Patrick~J
  Tighe, Azra Bihorac, et~al.
\newblock Intelligent icu for autonomous patient monitoring using pervasive
  sensing and deep learning.
\newblock {\em Scientific reports}, 9(1):1--13, 2019.

\bibitem{zhang2022transformer}
Anis Davoudi, Kumar~Rohit Malhotra, Benjamin Shickel, Scott Siegel, Seth
  Williams, Matthew Ruppert, Emel Bihorac, Tezcan Ozrazgat-Baslanti, Patrick~J
  Tighe, Azra Bihorac, et~al.
\newblock Transformer-based multimodal information fusion for facial expression analysis.
\newblock {\em Proceedings of the IEEE/CVF Conference on Computer Vision and Pattern Recognition}, pages 2428--2437, 2022.

\bibitem{meng2022valence}
Meng, Liyu and Liu, Yuchen and Liu, Xiaolong and Huang, Zhaopei and Jiang, Wenqiang and Zhang, Tenggan and Liu, Chuanhe and Jin, Qin.
\newblock Valence and arousal estimation based on multimodal temporal-aware features for videos in the wild.
\newblock {\em Proceedings of the IEEE/CVF Conference on Computer Vision and Pattern Recognition}, pages 2345--2352, 2022.

\bibitem{dosovitskiy2020image}
Dosovitskiy, Alexey and Beyer, Lucas and Kolesnikov, Alexander and Weissenborn, Dirk and Zhai, Xiaohua and Unterthiner, Thomas and Dehghani, Mostafa and Minderer, Matthias and Heigold, Georg and Gelly, Sylvain and others.
\newblock An image is worth 16x16 words: Transformers for image recognition at scale.
\newblock In {\em arXiv preprint arXiv}, 2010.11929, 2020.

\bibitem{liu2021swin}
Ze~Liu, Yutong Lin, Yue Cao, Han Hu, Yixuan Wei, Zheng Zhang, Stephen Lin, and
  Baining Guo.
\newblock Swin transformer: Hierarchical vision transformer using shifted
  windows.
\newblock In {\em Proceedings of the IEEE/CVF International Conference on
  Computer Vision}, pages 10012--10022, 2021.

\bibitem{kour2014real}
George Kour and Raid Saabne.
\newblock Real-time segmentation of on-line handwritten arabic script.
\newblock In {\em Frontiers in Handwriting Recognition (ICFHR), 2014 14th
  International Conference on}, pages 417--422. IEEE, 2014.

\bibitem{kour2014fast}
George Kour and Raid Saabne.
\newblock Fast classification of handwritten on-line arabic characters.
\newblock In {\em Soft Computing and Pattern Recognition (SoCPaR), 2014 6th
  International Conference of}, pages 312--318. IEEE, 2014.

\bibitem{hadash2018estimate}
Guy Hadash, Einat Kermany, Boaz Carmeli, Ofer Lavi, George Kour, and Alon
  Jacovi.
\newblock Estimate and replace: A novel approach to integrating deep neural
  networks with existing applications.
\newblock {\em arXiv preprint arXiv:1804.09028}, 2018.

\bibitem{kollias2023abaw2}
Dimitrios Kollias.
\newblock ABAW: Valence-Arousal Estimation, Expression Recognition, Action Unit Detection \& Emotional Reaction Intensity Estimation Challenges.
\newblock {\em arXiv preprint arXiv:2303.01498}, 2023.

\bibitem{kollias2023abaw}
Dimitrios Kollias.
\newblock ABAW: learning from synthetic data \& multi-task learning challenges.
\newblock {\em European Conference on Computer Vision}, pages 157--17, 2022.

\bibitem{kollias2022abaw}
Dimitrios Kollias.
\newblock Abaw: Valence-arousal estimation, expression recognition, action unit
  detection \& multi-task learning challenges.
\newblock {\em arXiv preprint arXiv:2202.10659}, 2022.

\bibitem{kollias2021analysing}
Dimitrios Kollias and Stefanos Zafeiriou.
\newblock Analysing affective behavior in the second abaw2 competition.
\newblock In {\em Proceedings of the IEEE/CVF International Conference on
  Computer Vision}, pages 3652--3660, 2021.

\bibitem{kollias2020analysing}
D~Kollias, A~Schulc, E~Hajiyev, and S~Zafeiriou.
\newblock Analysing affective behavior in the first abaw 2020 competition.
\newblock In {\em 2020 15th IEEE International Conference on Automatic Face and
  Gesture Recognition (FG 2020)(FG)}, pages 794--800.

\bibitem{kollias2021distribution}
Dimitrios Kollias, Viktoriia Sharmanska, and Stefanos Zafeiriou.
\newblock Distribution matching for heterogeneous multi-task learning: a
  large-scale face study.
\newblock {\em arXiv preprint arXiv:2105.03790}, 2021.

\bibitem{kollias2021affect}
Dimitrios Kollias and Stefanos Zafeiriou.
\newblock Affect analysis in-the-wild: Valence-arousal, expressions, action
  units and a unified framework.
\newblock {\em arXiv preprint arXiv:2103.15792}, 2021.

\bibitem{kollias2019expression}
Dimitrios Kollias and Stefanos Zafeiriou.
\newblock Expression, affect, action unit recognition: Aff-wild2, multi-task
  learning and arcface.
\newblock {\em arXiv preprint arXiv:1910.04855}, 2019.

\bibitem{kollias2019face}
Dimitrios Kollias, Viktoriia Sharmanska, and Stefanos Zafeiriou.
\newblock Face behavior a la carte: Expressions, affect and action units in a
  single network.
\newblock {\em arXiv preprint arXiv:1910.11111}, 2019.

\bibitem{kollias2019deep}
Dimitrios Kollias, Panagiotis Tzirakis, Mihalis~A Nicolaou, Athanasios
  Papaioannou, Guoying Zhao, Bj{\"o}rn Schuller, Irene Kotsia, and Stefanos
  Zafeiriou.
\newblock Deep affect prediction in-the-wild: Aff-wild database and challenge,
  deep architectures, and beyond.
\newblock {\em International Journal of Computer Vision}, pages 1--23, 2019.

\bibitem{zafeiriou2017aff}
Stefanos Zafeiriou, Dimitrios Kollias, Mihalis~A Nicolaou, Athanasios
  Papaioannou, Guoying Zhao, and Irene Kotsia.
\newblock Aff-wild: Valence and arousal ‘in-the-wild’challenge.
\newblock In {\em Computer Vision and Pattern Recognition Workshops (CVPRW),
  2017 IEEE Conference on}, pages 1980--1987. IEEE, 2017.

\end{thebibliography}
\end{document}